\documentclass[letterpaper, 10 pt, conference]{ieeeconf}  % Comment this line out if you need a4paper

\IEEEoverridecommandlockouts                              % This command is only needed if 
\usepackage{cite}
\usepackage{hyperref}
\usepackage[hyphenbreaks]{breakurl}
\usepackage{amsmath}
\usepackage{amssymb}
\usepackage{graphicx}
\usepackage{algorithmic}

\usepackage{algorithm,float}
\usepackage{lipsum}

\usepackage{multirow}
\usepackage{graphicx}

\makeatletter
\newenvironment{breakablealgorithm}
  {% \begin{breakablealgorithm}
   \begin{center}
     \refstepcounter{algorithm}% New algorithm
     \hrule height.8pt depth0pt \kern2pt% \@fs@pre for \@fs@ruled
     \renewcommand{\caption}[2][\relax]{% Make a new \caption
       {\raggedright\textbf{\ALG@name~\thealgorithm} ##2\par}%
       \ifx\relax##1\relax % #1 is \relax
         \addcontentsline{loa}{algorithm}{\protect\numberline{\thealgorithm}##2}%
       \else % #1 is not \relax
         \addcontentsline{loa}{algorithm}{\protect\numberline{\thealgorithm}##1}%
       \fi
       \kern2pt\hrule\kern2pt
     }
  }{% \end{breakablealgorithm}
     \kern2pt\hrule\relax% \@fs@post for \@fs@ruled
   \end{center}
  }
\makeatother

\title{\LARGE \bf
Goal-conditioned Hierarchical Reinforcement Learning for Sample-efficient and Safe Autonomous Driving at Intersections
}

\author{Yiou Huang$^{1}$% <-this % stops a space
\thanks{$^{1}$Yiou Huang
        {\tt\small leohyo009@163.com}}%
}

\begin{document}

\maketitle
\thispagestyle{empty}
\pagestyle{empty}

%%%%%%%%%%%%%%%%%%%%%%%%%%%%%%%%%%%%%%%%%%%%%%%%%%%%%%%%%%%%%%%%%%%%%%%%%%%%%%%%
\begin{abstract}
Reinforcement learning (RL) exhibits remarkable potential in addressing autonomous driving tasks. However, it is difficult to train a sample-efficient and safe policy in complex scenarios. In this article, we propose a novel hierarchical reinforcement learning (HRL) framework with a goal-conditioned collision prediction (GCCP) module. In the hierarchical structure, the GCCP module predicts collision risks according to different potential subgoals of the ego vehicle. A high-level decision-maker choose the best safe subgoal. A low-level motion-planner interacts with the environment according to the subgoal. Compared to traditional RL methods, our algorithm is more sample-efficient, since its hierarchical structure allows reusing the policies of subgoals across similar tasks for various navigation scenarios. In additional, 
the GCCP module's ability to predict both the ego vehicle's and surrounding vehicles' future actions according to different subgoals, ensures the safety of the ego vehicle throughout the decision-making process. Experimental results demonstrate that the proposed method converges to an optimal policy faster and achieves higher safety than traditional RL methods.

\end{abstract}

%%%%%%%%%%%%%%%%%%%%%%%%%%%%%%%%%%%%%%%%%%%%%%%%%%%%%%%%%%%%%%%%%%%%%%%%%%%%%%%%
\section{INTRODUCTION}
As autonomous driving technology advances, ensuring sample-efficient and safe navigation in complex urban environments remains a critical challenge \cite{alsharman2024autonomousdrivingunsignalizedintersections}. Interaction navigation, which encompasses the ability of autonomous vehicles to predict and respond to the behavior of surrounding agents plays a pivotal role in achieving seamless integration into mixed traffic scenarios. 

Rule-based methods, a cornerstone in autonomous systems development, use predefined heuristics and logic to govern vehicle behavior \cite{Kesting_2010}, \cite{6338774}, \cite{Bando_1998}. While offering predictable and interpretable outcomes, they often assume constant speed or acceleration, neglecting driver intentions and interactions between vehicles, especially at critical points like intersections. Additionally, their case-by-case mechanisms hinder generalization to new scenarios.

Imitation Learning (IL) has emerged as a promising alternative, enabling autonomous vehicles to learn from expert demonstrations. By mimicking the behavior of experienced drivers, these models can capture complex social interactions and adapt to varying contexts \cite{bojarski2016endendlearningselfdriving}, \cite{bansal2018chauffeurnetlearningdriveimitating}.
However, imitation learning faces notable limitations due to its reliance on high-quality training data and difficulty generalizing to unseen situations. Additionally, minor errors can quickly accumulate, causing the agent to deviate from the expert's intended path—especially in environments that differ slightly from training scenarios. This issue, known as "covariate shift," can greatly impact performance.

Reinforcement Learning (RL) introduces a more adaptive approach, allowing vehicles to learn optimal navigation strategies through trial and error \cite{isele2018navigatingoccludedintersectionsautonomous}, \cite{tram2018learningnegotiatingbehaviorcars},
\cite{leurent2019socialattentionautonomousdecisionmaking},
\cite{9304542}. By maximizing cumulative rewards based on interactions with their environment, RL algorithms can develop sophisticated policies that enhance decision-making in uncertain and dynamic contexts. Nonetheless, the sample inefficiency and safety concerns associated with exploration in real-world environments pose substantial challenges. Recently, numerous studies have explored strategies to improve sample efficiency through Hierarchical Reinforcement Learning (HRL) \cite{8569400}, \cite{qiao2020behaviorplanningurbanintersections}, \cite{lu2023actiontrajectoryplanningurban}. 
However, the high-level decision-making in these methods lacks security assessment, making their implementation in real-world scenarios hazardous.
Some studies have focused on enhancing safety by predicting the internal states or behavioral intentions of surrounding traffic vehicles \cite{8593420}, \cite{liu2022learningnavigateintersectionsunsupervised}, \cite{10375253}, \cite{10530461}. However, these methods are constrained to predicting the most likely intentions of surrounding vehicles and fail to adapt to the diverse potential future plans of the ego vehicle.
In this paper, we propose a multi-task HRL framework to enhance sample efficiency in learning policies within complex interactive environments. Additionally, we introduce a Goal-Conditioned Collision Prediction (GCCP) module that simulates the reactions of surrounding vehicles based on different subgoals of the ego vehicle, ensuring safe decision-making.
The main contributions of this paper are as follows.
\begin{itemize}
    \item Proposes a multi-task HRL framework integrated with a GCCP module, enhancing sample efficiency and safety while improving the interpretability of interactions between the ego vehicle and surrounding vehicles.
    \item Proposes a novel GCCP module aimed at preventing collisions in intersection navigation scenarios.
    \item Evaluates the methods through a series of experiments in the SMARTS simulator, demonstrating that the proposed approach outperforms traditional RL methods.
\end{itemize}

\begin{figure}[!t]
    \centering
    \includegraphics[width=8.5cm]{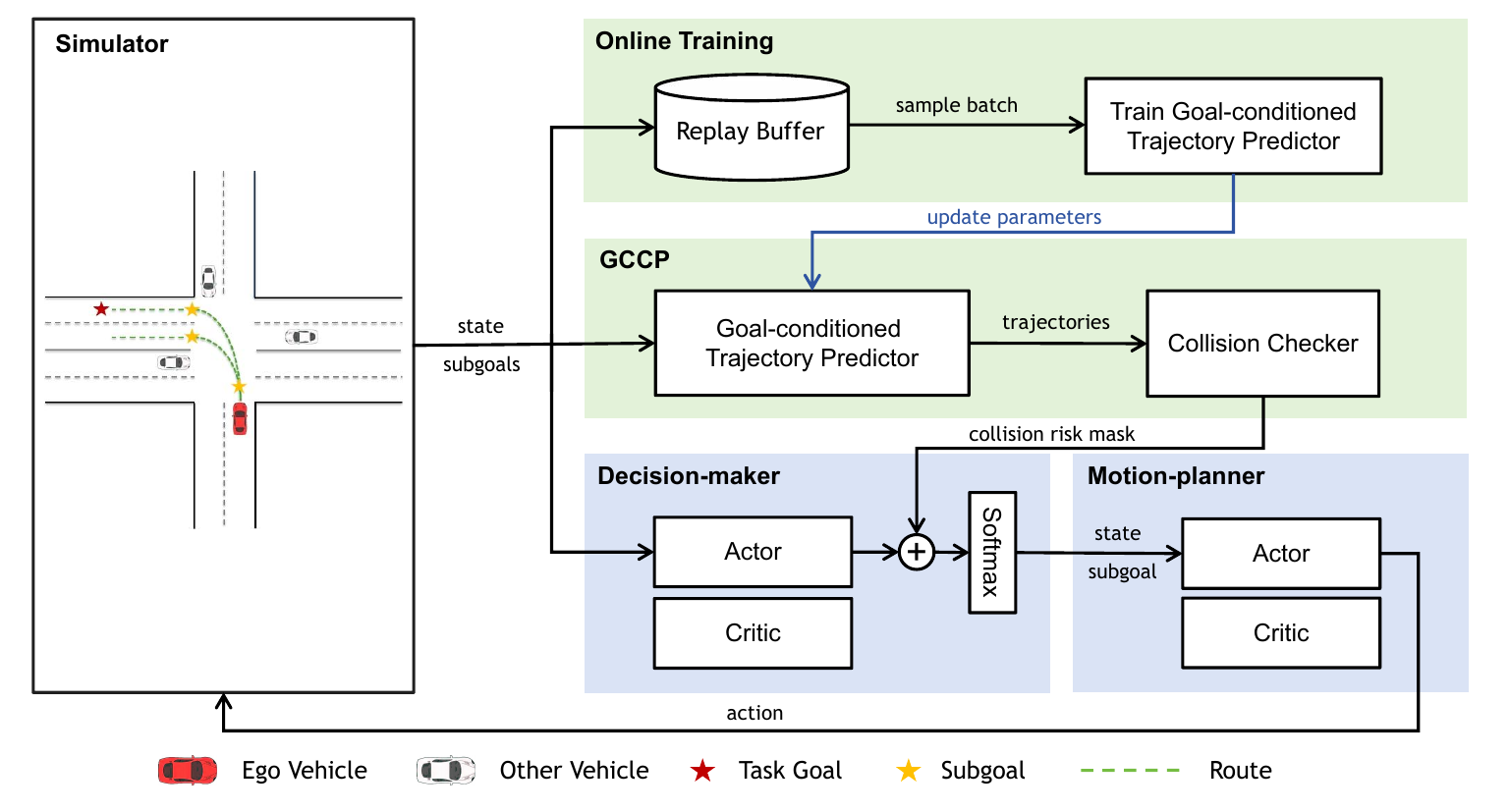}
    \caption{An overview of the framework. The GCCP module outputs the collision risk mask, which the decision-maker uses to select the best subgoal. The motion-planner interacts directly with the environment to approach the selected subgoal. Episodic data is stored in a replay buffer, and batch data is sampled from it to train the goal-conditioned trajectory predictor.}
    \label{fig1}
\end{figure}

\section{RELATED WORKS}

\subsection{Hierarchical Reinforcement Learning}
HRL has emerged as a key approach to address scalability challenges in traditional RL for complex, high-dimensional tasks. By decomposing policies into multiple levels, HRL enables higher-level policies to set subgoals for lower-level policies, thereby reducing the exploration space and enhancing learning efficiency. The foundational work by \cite{SUTTON1999181} introduces "options" as temporally extended actions, while \cite{NIPS1992_d14220ee} proposes a feudal architecture where high-level managers assign subgoals to lower-level workers. Recent advancements have integrated deep learning techniques into HRL frameworks. For instance, \cite{kulkarni2016hierarchicaldeepreinforcementlearning} developed the hierarchical-DQN (h-DQN), which employs high-level deep Q-networks (DQNs) to select subgoals and low-level DQNs to execute actions. To address the issue of non-stationarity—where the learning dynamics of one level affect the stability of others—\cite{nachum2018dataefficienthierarchicalreinforcementlearning} introduced an off-policy correction method. Additionally, \cite{levy2019learningmultilevelhierarchieshindsight} proposed a training approach where each level is optimized under the assumption that lower levels are already optimal, further improving the robustness of multi-level policy learning.

\subsection{Trajectory Prediction}
Trajectory prediction in autonomous driving has garnered significant attention due to its critical role in enhancing safety and decision-making processes. Recent advancements in deep learning and probabilistic modeling have led to the development of various approaches for predicting the trajectories of surrounding vehicles. 
% \cite{Deo_2018} utilizes Recurrent Neural Network (RNN) to model vehicle trajectories and predict future positions.
\cite{chai2019multipathmultipleprobabilisticanchor} employs a Convolutional Neural Network (CNN) to extract mid-level features, predicting discrete distributions over fixed trajectory anchors.
\cite{liang2020learninglanegraphrepresentations} models the interactions among lanes-lanes, lanes-agents and agents-agents with Graph Convolutional Network (GCN) and attention-based mechanisms. 
\cite{zhao2020tnttargetdriventrajectoryprediction} predicts future trajectories by generating and refining target points based on driving context and agent interactions.
However, these methods do not model surrounding vehicles' reactions to various future decisions of the ego vehicle. 
\cite{Song_2020}, \cite{huang2023learninginteractionawaremotionprediction} predict other vehicles' future trajectories based on the ego vehicle's planned future trajectories. Nevertheless, these planned trajectories are human-designed and could not scale to complex situations. \cite{ngiam2022scenetransformerunifiedarchitecture} effectively captures interactions between multiple agents in a scene, enabling prediction of agents' behavior conditioned on the goal of the ego vehicle.
Inspired by \cite{ngiam2022scenetransformerunifiedarchitecture}, we utilize a goal-conditioned trajectory predictor in our GCCP module. Furthermore, we adopt an online training framework rather than offline training, allowing seamless cooperation with our hierarchical decision-maker and motion-planner.

\subsection{Intersection Navigation}
Navigating intersections in dense traffic is one of the most complex challenges faced by autonomous vehicles in urban environments. To address this issue, a range of RL methods have been proposed. 
\cite{isele2018navigatingoccludedintersectionsautonomous} examines the potential of deep reinforcement learning to tackle intersection navigation issues.
\cite{tram2018learningnegotiatingbehaviorcars} learns a policy to select a high-level action that adjusts the ego vehicle's speed along its pre-defined trajectory to navigate the crossing.
\cite{leurent2019socialattentionautonomousdecisionmaking} introduces an attention-based architecture designed to model the interaction patterns among a dynamic number of nearby vehicles.
\cite{9304542} propses a unified learning framework and introduces a vectorized reward function integrated with DQN to simultaneously learn and manage multiple intersection navigation tasks.
Recently, several studies have focused on enhancing sample efficiency through HRL.
\cite{8569400} formulates the intersection navigation problem as a Partially Observable Markov Decision Process (POMDP) and generates both high-level discrete options and low-level actions to address it.
\cite{qiao2020behaviorplanningurbanintersections} introduces a HRL framework with multiple layers dedicated to behavior planning at urban intersections.
\cite{lu2023actiontrajectoryplanningurban} employs two off-policy actor-critic structures to train decision layers for trajectory generation.
Several studies have focused on enhancing safety by predicting the internal traits or intentions of surrounding traffic vehicles.
\cite{8593420} uses a Kalman filter to estimate surrounding vehicle positions, aiding the ego vehicle in avoiding hazardous decisions.
\cite{liu2022learningnavigateintersectionsunsupervised} presents an unsupervised approach for learning representations of driver traits to improve autonomous vehicle navigation at uncontrolled T-intersections.
\cite{10375253} uses the Social Value Orientation (SVO) in the decision-making of vehicles to describe the social interaction among multiple vehicles. 
% \cite{10530461} presents a reinforcement learning approach for interactive autonomous navigation, incorporating three auxiliary tasks: internal state inference, trajectory prediction, and interactivity estimation.
\cite{10530461} proposes a reinforcement learning approach for interactive autonomous navigation, integrating auxiliary tasks for internal state inference, trajectory prediction, and interactivity estimation.

\section{METHOD}
We introduce a hierarchical framework that can efficiently learn a hierarchical policy to perform intersection navigation tasks, which contains two components: (1) a GCCP module (2) a hierarchical decision-maker and motion-planner. Fig. \ref{fig1} illustrates the architecture of the framework.

\subsection{Problem Formulation}

In this work, intersection navigation is formulated as a Markov Decision Process (MDP) to enable efficient and safe decision-making for autonomous driving. An MDP is defined by the tuple $(\mathcal{S}, \mathcal{A}, \mathcal{P}, \mathcal{R}, \gamma)$, where $\mathcal{S}$ represents the state space, $\mathcal{A}$ is the action space, $\mathcal{P}: \mathcal{S} \times \mathcal{A} \times \mathcal{S} \rightarrow [0, 1]$ denotes the state transition probability, $\mathcal{R}: \mathcal{S} \times \mathcal{A} \rightarrow \mathbb{R}$ is the reward function, and $\gamma \in [0, 1]$ is the discount factor. The goal is to find an optimal policy that maps states to actions to maximize the expected cumulative reward.

\subsection{State Space and Subgoals}

\subsubsection{State Space}
\label{s1}
The environment state $s$ contains five components: ego vehicle's history trajectory, surrounding vehicles' history trajectories, ego vehicle's routes, surrounding vehicles' routes and drivable area. 

For history trajectories and routes, we implement a vectorized representation strategy known as VectorNet \cite{DBLP:journals/corr/abs-2005-04259} to enhance computational and memory efficiency. 
For each surrounding vehicle $i$, the historical trajectory is defined as $\chi^{i}_{-T_h:0}=\left \{ v^{i}_{-T_h}, v^{i}_{-T_h+1}, ..., v^{i}_{j}, ..., v^{i}_{0} \right \}$, where $i \in [0, N_s]$, and $N_s$ is the number of surrounding vehicles. Each element $v^{i}_{j}=\left [ p^{i}_{j-1}, p^{i}_{j}, i \right ]$ represents the trajectory vector between two consecutive poses, where $p^{i}_{j}$ is the pose of vehicle $i$ at timestamp $j$, consisting of $(x, y, heading, speed)$.  
The planned routes for vehicle $i$ are given by $\chi^{i}_{r}=\left \{ r^{i}_{0}, r^{i}_{1}, ..., r^{i}_{j}, ..., r^{i}_{N_r-1} \right \}$, where $N_r$ is the number of routes. Each route $r^{i}_{j}$ contains a sequence of waypoint vectors $r^{i}_{j}=\left \{ v^{j}_{0}, v^{j}_{1}, ..., v^{j}_{k}, ..., v^{j}_{N_p-1} \right \}$, with $N_p$ representing the number of waypoints. Each element $v^{j}_{k}=\left [ p^{j}_{k}, p^{j}_{k+1}, i, j \right ]$ encodes information between consecutive waypoints, where $p^{j}_{k}=(x, y, heading)$ is the $k$-th waypoint of route $j$.  
We also include the ego vehicle's historical trajectory $\chi^{e}_{-T_h:0}$ and planned routes $\chi^{e}_{r}$. All elements are represented in the ego vehicle's coordinate system. In our experiments, we consider $N_s=5$ surrounding vehicles closest to the ego vehicle. For both the ego and surrounding vehicles, we select $N_r=3$ routes toward the navigation task goal, each containing $N_p=50$ waypoints spaced at 1-meter intervals. Historical trajectories and routes are padded with zeros if there are fewer surrounding vehicles or insufficient historical data.  

For the drivable area, we represent the state using a $64 \times 64 \times 1$ Bird's-Eye View (BEV) image $I_{d}$. The resolution maps 64 pixels to a 50-meter span, with each pixel assigned a value: 0 for undrivable areas and 1 for drivable areas. Rendering is based on the position of the ego vehicle. The ego vehicle is placed at the center of the image.

\subsubsection{Subgoals} 
\label{subgoal}
The potential subgoals form a set of $N$ points sampled from the ego vehicle's routes: $\mathcal{G} = \left \{ g^{i} \right \} = \left \{ \left ( x^{i}, y^{i}, heading^{i} \right ) \right \}^{N-1}_{i=0}$. In practice, we sample $N=12$ subgoals within a 20-meter range, maintaining a 5-meter distance between adjacent points. If fewer than 12 points are available along the routes, the remaining subgoals are filled using the sampled subgoal closest to the task goal.

\subsection{Goal-conditioned Collision Prediction}
This module predicts potential future scenes for the ego vehicle's subgoals and identifies collision risks. It comprises two components: (1) a neural network-based goal-conditioned trajectory predictor and (2) a rule-based collision checker.
\subsubsection{Goal-conditioned Trajectory Predictor}
\label{gctr}
\begin{figure}[!t]
    \centering
    \includegraphics[width=8.5cm]{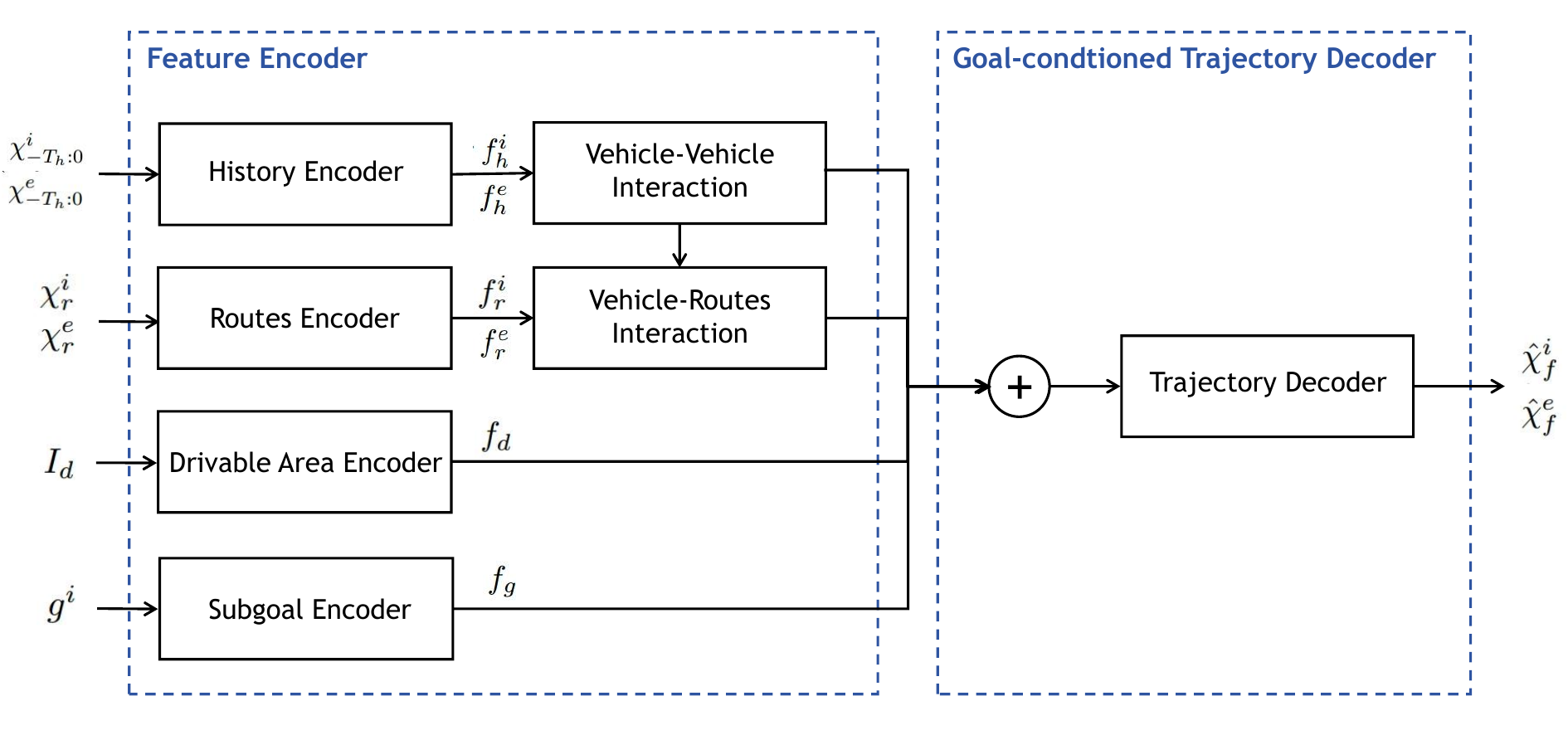}
    \caption{An overview of the goal-conditioned trajectory predictor: The vehicles' history, planned routes, drivable area, and the ego vehicle's subgoal are encoded using a feature encoder. A trajectory decoder then predicts the future trajectories of both the ego vehicle and surrounding vehicles, conditioned on the ego vehicle's subgoal.}
    \label{fig2}
\end{figure}

The predictor generates future trajectories for the ego and surrounding vehicles conditioned on each subgoal. Fig. \ref{fig2} illustrates the architecture of the goal-conditioned trajectory predictor.

\textbf{Feature Encoder.}
Each surrounding vehicle $i$'s history trajectory $\chi^{i}_{-T_h:0}$ is encoded into a fixed-length $L = 128$ embedding vector $f^{i}_{h}$ through a shared self-attention \cite{DBLP:journals/corr/VaswaniSPUJGKP17} history encoder. The ego vehicle's history trajectory $\chi^{e}_{-T_h:0}$ is also encoded into a fixed-length $L$ embedding vector $f^{e}_{h}$ through the same history encoder. Both the ego vehicle's and surrounding vehicles' embedding vectors are gathered as nodes in the interaction graph, and a self-attention module is used to process the graph to capture the relationship between the interacting vehicles. Similarly, each surrounding vehicle $i$'s routes $\chi^{i}_{r}$ is encoded into fixed-length $L$ embedding vector $f^{i}_{r}$ through a shared self-attention route encoder. The ego vehicle's routes $\chi^{e}_{r}$ is also encoded into fixed-length $L$ embedding vector $f^{e}_{r}$ through the same route encoder. Then, we utilize a cross-attention layer with the query as the interaction feature vector of the vehicle and key and value as the feature vectors of the vehicle's routes, to capture the relationship between the vehicle and routes. The drivable area is encoded into a fixed-length $L$ embedding vector $f_{d}$ using a CNN. For subgoal feature, we use a 2-layer Multi-Layer Perception (MLP) to encode into a fixed-length $L$ embedding vector $f_{g}$. Finally, for each vehicle, its historical feature, interaction feature, route attention feature, drivable area feature and subgoal feature are concatenated and passed through the goal-conditioned trajectory decoder to generate the predicted future trajectory.

\textbf{Goal-conditioned Trajectory Decoder.}
This stage uses a 2-layer MLP to predict future trajectories for each surrounding vehicle $\hat{\chi}^{i}_{f} = \{\hat{p}^{i}_{1}, \hat{p}^{i}_{2}, ..., \hat{p}^{i}_{T_f}\}$ and the ego vehicle $\hat{\chi}^{e}_{f} = \{\hat{p}^{e}_{1}, \hat{p}^{e}_{2}, ..., \hat{p}^{e}_{T_f}\}$, where $\hat{p}$ represents $(x, y, heading)$. To ensure interaction-aware predictions, the module incorporates the ego vehicle's subgoal to model its influence on surrounding vehicles. The prediction horizon is 1 second ($T_f=10$) with a 0.1-second time interval.

\subsubsection{Collision Checker}
\label{masking}
The collision checker takes as input the future trajectories of the ego and surrounding vehicles, conditioned on $N$ subgoals from the trajectory predictor. It outputs a collision risk mask $m^{c} = \left [ c_{0}, c_{1}, \dots, c_{i}, \dots, c_{N-1} \right ]$, where
\begin{equation}
c_{i} =
\begin{cases}
-10^{8} & \text{if collision,}  \\
0 & \text{otherwise}
\end{cases}
\end{equation}
indicating whether the ego vehicle's predicted trajectory collides with the predicted trajectories of surrounding vehicles under subgoal $g^{i}$.
 The collision condition is evaluated using the Separating Axis Theorem (SAT) \cite{Huynh2009SeparatingAT}.

\subsection{Hierarchical Decision-making and Motion-planning}
\begin{figure}[!t]
    \centering
    \includegraphics[width=8.5cm]{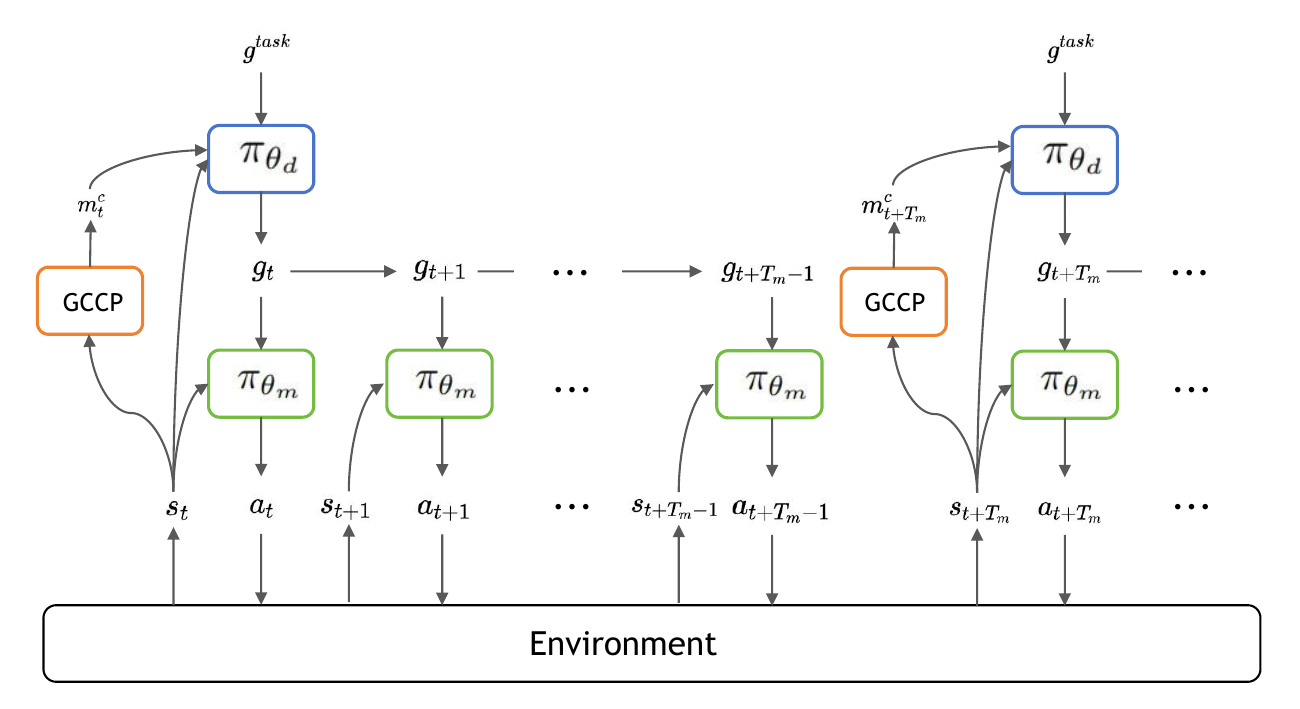}
    \caption{The design of the hierarchical decision-making and motion-planning. The lower-level motion-planner's policy interacts directly with the environment. The high-level decision-maker's policy observes the environment state, the task goal and the collision risk mask provided by the GCCP module and instructs the motion-planner via subgoals, which are sampled anew every $T_m$ steps.
    }
    \label{fig3}
\end{figure}

The hierarchical two-layer policy consists of a high-level decision-maker $\pi_{\theta_{d}}$ and a low-level motion-planner $\pi_{\theta_{m}}$ (see Fig. \ref{fig3}). The decision-maker sets subgoal for the motion-planner, corresponding to state the ego vehicle aims to reach. At each timestep $t$ (when $t \mod T_m = 0$, $T_m$ being the motion-planner's action steps), the decision-maker observes the environment state $s_t$, task goal $g^{\text{task}}$, and collision risk mask $m^c_t$ from the GCCP module, sampling a subgoal $g_t$. The motion-planner observes $s_t$ and $g_t$, generating an action $a_t$ applied to the environment, which yields a reward $r^m_t$ and transitions to $s_{t+1}$. After $T_m$ steps, the decision-maker's reward $r^d_t$ is computed. To align planning and prediction horizons, $T_m = T_f = 10$.

\subsubsection{State Space for Decision-maker}
The decision-maker's state $s^{d}_{t} = (s_{t} , g^{task})$ at time step t consists of two parts: an environment state $s_{t}$ and a task goal $g^{task}$. The task goal can be repesented as $(x^{task},y^{task},heading^{task})$.

\subsubsection{Action Space for Decision-maker}
The decision-maker's action space $\mathcal{A}^{d}_{t}$ is the same as the subgoal sets $\mathcal{G}_{t}$ introduced in \ref{subgoal}.

\subsubsection{Unsafe Action Masking for Decision-maker}
The decision-maker's policy outputs logits \(l^{d}_t = [l^d_{t,0}, l^{d}_{t,1}, ..., l^{d}_{t,N-1}]\), which are converted into an action probability distribution using a softmax operation. To prevent unsafe subgoals, a collision risk mask \(m^c_t\), provided by the GCCP module, is applied by adding large negative values to the logits of unsafe actions. The re-normalized probability distribution \(\pi_{\theta_d}(g_t|s^{d}_{t}, m^{c}_{t})\) is computed as:
\begin{equation}
\pi_{\theta_d}(g_t|s^{d}_{t}, m^{c}_{t}) = \text{softmax}(l^{d}_t + m^c_{t}),
\end{equation}
ensuring that the probability of selecting unsafe subgoals is effectively zero.

\subsubsection{Neural Network for Decision-maker}
In practice, we choose PPO \cite{DBLP:journals/corr/SchulmanWDRK17} as the algorithm. 
The module consists of a policy $\pi_{\theta_d}$ and a value function $V_{\phi_d}$.
Their feature encoders are similar with that of the goal-conditioned trajectory predictor. The feature encoders extract the state feature from the environment state and task goal. We use another 2-layer MLP to encode $N$ subgoals. In the policy decoder, for each subgoal, we concatenate the state feature with the corresponding subgoal feature. The concatenated feature is then passed through a 3-layer MLP, which outputs a one-dimensional logit. All $N$ logits are concatenated and processed through a softmax function to obtain the probabilities of the subgoals.
In the value function decoder, we similarly concatenate the state feature with each subgoal feature. The concatenated features are fed into a 2-layer MLP. For each subgoal, this produces a hidden feature of dimension $L$. These hidden features are concatenated, flattened into a vector, and passed through another 2-layer MLP to generate a one-dimensional value.

\subsubsection{Reward Function for Decision-maker}
The reward function of the decision-maker is defined as follows:
\begin{equation}
\label{eqn_rd}
r^{d}_t = r^{goal}_t + r^{subgoal}_t + r^{subgoal\_dist}_t
\end{equation}
In particular, $r^{goal}_{t}$ is the reward when the ego vehicle reaches the task goal:
\begin{equation}
r^{goal}_{t} =
\begin{cases}
r_{goal} & \text{if task goal is reached,} \\
0 & \text{otherwise}
\end{cases}
\end{equation}

$r^{subgoal}_t$ is the reward when the ego vehicle reaches the subgoal in $T_m$ steps.
\begin{equation}
r^{subgoal}_{t} =
\begin{cases}
r_{subgoal} & \text{if subgoal is reached,} \\
0 & \text{otherwise}
\end{cases}
\end{equation}

$r^{subgoal\_dist}_t$ represents the reward designed to encourage the decision-maker to select a subgoal located far from the ego vehicle.
\begin{equation}
r^{subgoal\_dist}_{t} = r_{near} + {\varepsilon}_s \times d_{subgoal} 
\end{equation}
where $r_{near}$ is a constant, ${\varepsilon}_s$ is a hyper-parameter, $d_{subgoal} $ is the distance between the ego vehicle and the subgoal.
In our experiments, we set $r_{goal} = 3$, $r_{subgoal} = 0.5$, $r_{near}=-0.5$, ${\varepsilon}_s=0.05$.

\subsubsection{State Space for Motion-planner}
The motion-planner's state $s^{m}_{t} = (s_{t} , g_t)$ at time step t consists of two parts: an environment state $s_{t}$ and a subgoal $g_t$ provided by the decision-maker.

\subsubsection{Action Space for Motion-planner}
The action space is discrete and consists of a set of $N_a$ actions, $\mathcal{A}_m = \left \{ a^i\right \} = \left \{ \left ( x_{delta}^i, y_{delta}^i, heading_{delta}^i \right )\right\}^{N_a-1}_{i=0}$, where $x_{delta}$, $y_{delta}$ and $heading_{delta}$ represent the changes in the x-coordinate, y-coordinate, and heading, respectively, to be reached in 0.1 seconds. In practice, we design $N_a=6$ actions: slow down, keep direction, slowly turn left, slowly turn right, quickly turn left, and quickly turn right.

\subsubsection{Neural Network for Motion-planner}
In practice, we choose PPO \cite{DBLP:journals/corr/SchulmanWDRK17} as the algorithm. It contains a policy $\pi_{\theta_m}$ and a value function $V_{\phi_m}$. 
Their feature encoders are identical to that of the goal-conditioned trajectory predictor. The feature encoders extract state features from the environment state and the subgoal. In the policy decoder, the state feature is processed by a 3-layer MLP to produce an $N_a$-dimensional probability vector. Similarly, in the value function decoder, the state feature is passed through a 3-layer MLP to generate a one-dimensional value.

\subsubsection{Reward Function for Motion-planner}
The reward function of the motion-planner is defined as follows:
\begin{equation}
\label{eqn_rm}
r^{m}_t = r^{time}_t + r^{arr}_t + r^{col}_t + r^{off}_t + r^{dist}_t + r^{head}_t
\end{equation}

In particular, $r^{time}_t$ is a penalty for time, it is a constant.

$r^{arr}_t$ is the reward when the ego vehicle reaches the subgoal.
\begin{equation}
r^{arr}_{t} =
\begin{cases}
r_{arr} & \text{if subgoal is reached,} \\
0 & \text{otherwise}
\end{cases}
\end{equation}

$r^{col}_t$ is the penalty when the ego vehicle collides with other vehicles.
\begin{equation}
r^{col}_{t} =
\begin{cases}
r_{col} & \text{if collsion,} \\
0 & \text{otherwise}
\end{cases}
\end{equation}

$r^{off}_t$ represents the penalty incurred when the ego vehicle goes off the road.
\begin{equation}
r^{off}_{t} =
\begin{cases}
r_{off} & \text{if go off road,} \\
0 & \text{otherwise}
\end{cases}
\end{equation}

We use $r^{dist}_t$ to encourage the ego vehicle to move closer to the subgoal's position compared to the previous timestep.
\begin{equation}
r^{dist}_{t} = {\varepsilon}_d \times \left ( d^{s}_{t-1} - d^{s}_{t} \right )
\end{equation}
where ${\varepsilon}_d$ is a hyper-parameter, $d^{s}_{t-1}$ represents the distance between the ego vehicle and the subgoal at the previous timestep, $d^{s}_{t}$ represents the distance between the ego vehicle and the subgoal at current timestep.

We use $r^{head}_t$ is used to encourage the ego vehicle to decrease the orientation difference with the subgoal compared to the previous timestep.
\begin{equation}
r^{head}_{t} = {\varepsilon}_h \times \left ( h^{s}_{t-1} - h^{s}_{t} \right )
\end{equation}
where ${\varepsilon}_h$ is a hyper-parameter, $h^{s}_{t-1}$ represents the orientation difference between the ego vehicle and the subgoal at the previous timestep, $h^{s}_{t}$ represents the orientation distance between the ego vehicle and the subgoal at current timestep. 
In our experiments, we set $r^{time}_t = -0.05$, $r_{arr} = 1.0$, $r_{col}=-1$, $r_{off}=-1$, ${\varepsilon}_d=0.05$, ${\varepsilon}_h=0.5$.

\subsection{Training Framework}
We develop an online learning procedure for the framework, as outlined in Algorithm \ref{alg1}. At each timestep \( t \) where \( t \mod T_m = 0 \), the GCCP module outputs a collision risk mask. The decision-maker uses the current state, task goal, and collision risk mask to select a subgoal, guiding the motion-planner to generate an action. The environment responds with a reward and transitions to a new state. The motion-planner stores the experience \((s_t, g_t, a_t, r^m_t, s_{t+1})\) in buffer \(\mathcal{D}_m\). After \(T_m\) steps, the motion-planner's policy and value function are trained using \(\mathcal{D}_m\) with \(K_m\) gradient steps, after which \(\mathcal{D}_m\) is cleared. Simultaneously, every \(T_m\) steps, the decision-maker stores \((s_t, g^{task}, m^c_t, g_t, r^d_t, s_{t+T_m})\) in buffer \(\mathcal{D}_d\). At the end of each episode, the decision-maker is trained using \(\mathcal{D}_d\) with \(K_d\) gradient steps, and \(\mathcal{D}_d\) is cleared. 
Additionally, an episode buffer \(\mathcal{D}_e\) stores environment states, decision-maker-selected subgoals, and all vehicle trajectories, which are transferred to a replay buffer \(\mathcal{D}\) at the end of each episode.

Inspired by \cite{huang2023learninginteractionawaremotionprediction}, we train the goal-conditioned trajectory predictor at the end of each episode with $K_p$ gradient steps. At each gradient step, we sample a batch of data from the replay buffer $\mathcal{D}$, ensuring that the timestep $t$ of each data point satisfies $t \mod T_m = 0$.
The batch size is $N_b$ and the loss function for the batch is:
\begin{equation}
\begin{aligned}
\label{eq2}
\ell  = \frac{1}{N_b (N_s+1)}\sum_{i=0}^{N_b-1} (\ell_{SL1}(\hat{\chi}^e_f, \chi^e_f) + \sum_{j=0}^{N_s-1} \ell_{SL1}(\hat{\chi}^j_f, \chi^j_f) )
\end{aligned}
\end{equation}
where $\ell_{SL1}$ is the smooth L1 loss, $\chi^e_f$ is the ego vehicle's ground truth trajectory and $\chi^j_f$ is the $j$-th surrounding vehicle's ground truth trajectory.

Since the motion-planner's policy converges quickly, and we aim for the goal-conditioned trajectory predictor to learn a stable ego vehicle trajectory, we stop updating the motion-planner after the $N_g$-th episode and begin training the goal-conditioned trajectory predictor at the same time. To ensure the collision risk mask provided to the decision-maker is reliable, the output of the GCCP module is only passed to the decision-maker after the $N_m$-th episode. Until then, the collision risk mask is set to an $N$-dimensional zero vector. We train for a total of $N_e$ episodes.
In our experiments, we set $K_m=5$, $K_d=5$, $K_p=5$, $N_b=64$, $N_e=2000$, $N_g=500$, $N_m=800$.

\begin{breakablealgorithm}
    \caption{Training Framework}
    \label{alg1}
    \renewcommand{\algorithmicrequire}{\textbf{Input:}}

    \begin{algorithmic}[1]
        \REQUIRE episodic buffer $\mathcal{D}_e$, replay buffer $\mathcal{D}$, goal-conditioned trajectory predictor network $\theta$, 
                 decision-maker policy network $\pi_{\theta_d}$,
                 decision-maker value function network $V_{\phi_d}$,
                 decision-maker buffer $\mathcal{D}_d$,
                 motion-planner policy network $\pi_{\theta_m}$,
                 motion-planner value function network $V_{\phi_m}$,
                 motion-planner buffer $\mathcal{D}_m$
                 
        \FOR{e $\leftarrow$ 1 to $N_e$}
            \STATE Reset environment state
            \STATE Empty the episodic buffer $\mathcal{D}_e$
            \STATE Set t = 0
            \WHILE{not terminate}
                \STATE $m^c_{t}$ = $\overrightarrow{0}$
                \IF{ e $\textgreater$ $N_m$} 
                    \STATE Get collision risk mask $m^c_{t}$ from GCCP module
                \ENDIF
                \STATE Sample subgoal $g_t \sim \pi_{\theta_d}(g_t|s_t, g^{task}, m^c_{t})$

                \FOR{$t_m$ = 0 to $T_m-1$}
                    \STATE Sample motion-planner action $a_{t+t_m} \sim \pi_{\theta_m}(a_{t+t_m} |s_{t+t_m} , g_t)$
                    \STATE Execute $a_{t+t_m}$, observe next state $s_{t+t_m+1}$ and compute motion-planner reward $r^{m}_{t+t_m}$
                    \STATE Store $(s_{t+t_m}, g_{t}, a_{t+t_m}, r^{m}_{t+t_m}, s_{t+t_m+1})$ to $\mathcal{D}_m$
                \ENDFOR

                \STATE Compute decision-maker reward $r^{d}_{t}$
                \STATE Store $(s_{t}, g^{task}, m^c_{t}, g_{t}, r^{d}_{t},  s_{t+t_m+1})$ to $\mathcal{D}_d$

                \IF{ e $\textless$ $N_g$} 
                    \STATE Update $\theta_m$ and $\phi_m$ according to \cite{DBLP:journals/corr/SchulmanWDRK17} with $K_m$ gradient steps 
                \ENDIF
                \STATE Empty the buffer $\mathcal{D}_m$

                \STATE $t = t + t_m + 1$

            \ENDWHILE
            
            \STATE Update $\theta_d$ and $\phi_d$ according to \cite{DBLP:journals/corr/SchulmanWDRK17} with $K_d$ gradient steps
            \STATE Empty the buffer $\mathcal{D}_d$
            
            \IF{ e $\textgreater$ $N_g$} 
                \STATE Dump episodic buffer $\mathcal{D}_e$ into replay buffer $\mathcal{D}$
                \FOR{Step $\leftarrow$ 1 to $K_p$}
                    \STATE Initialize a mini-batch $\mathcal{B}$
                    \STATE Sample $N_b$ data from replay buffer $\mathcal{D}$ and add to mini-batch $\mathcal{B}$
                    \STATE Calculate loss according to Eq.\ref{eq2}
                    \STATE Update the model parameter $\theta$
                \ENDFOR
            \ENDIF
        \ENDFOR
        
    \end{algorithmic}
\end{breakablealgorithm}

\section{EXPERIMENTS}
\begin{figure}[!t]
    \centering
    \includegraphics[width=8.5cm]{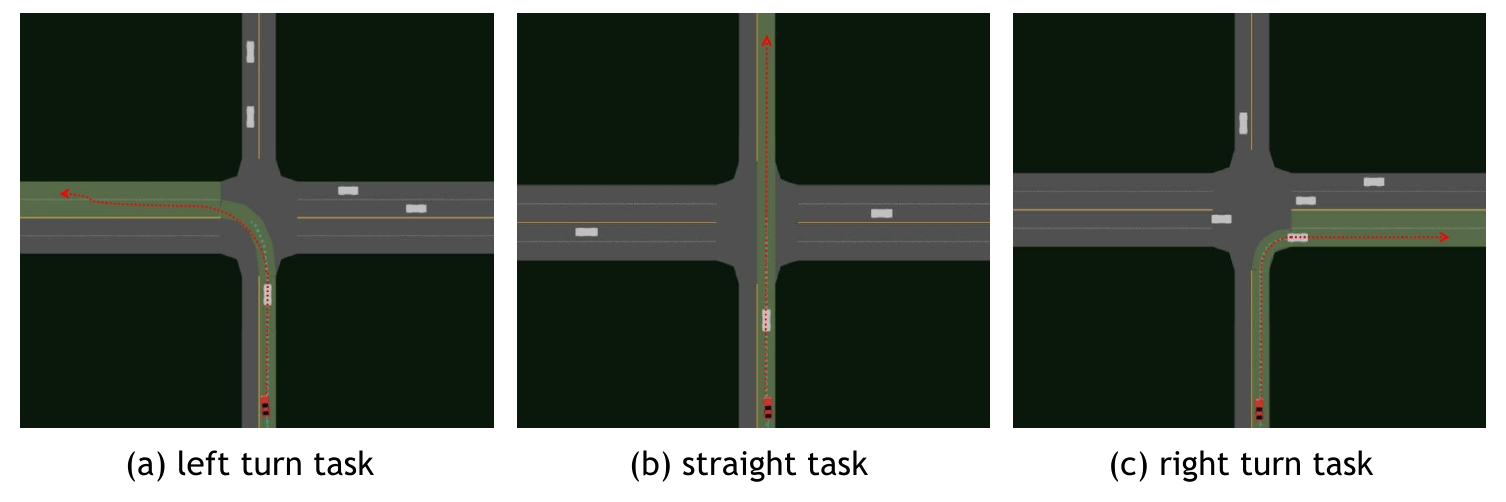}
    \caption{Three designed intersection scenarios in the SMARTS simulator}
    \label{fig4}
\end{figure}

In this paper, we evaluate the performance of the proposed method to deal with three intersection navigation scenarios in the SMARTS simulator \cite{zhou2020smartsscalablemultiagentreinforcement}, which are displayed in Fig. \ref{fig4}. The SMARTS simulator is an open-source, scalable platform designed for developing and evaluating multi-agent reinforcement learning algorithms in autonomous driving scenarios.

\subsection{Experimental Setup}
\subsubsection{Baseline methods}
To evaluate the performance of our proposed framework, we established several baseline methods. We utilized traditional RL algorithms DQN \cite{DBLP:journals/corr/MnihKSGAWR13}, PPO \cite{DBLP:journals/corr/SchulmanWDRK17} and SAC-Discrete \cite{christodoulou2019softactorcriticdiscreteaction} to train the agent. The neural networks utilized for the policy and value function are identical to those employed in the motion-planner. The definitions of state, action, and reward remain consistent with those used in the motion-planner within our framework, with the subgoal being replaced by the task goal.

\subsubsection{ablated methods}
Three sub-models are constructed by removing key components of the framework to evaluate their individual functions. First, the GCCP module is removed to assess its role in enhancing safety. Second, the subgoal feature is replaced with a zero vector when predicting the trajectories of surrounding vehicles, implying that other vehicles do not respond to the ego vehicle's subgoal. Third, a constant velocity (CV) model is used to predict the trajectories of surrounding vehicles instead of a learning-based predictor.

\subsection{Implementation Details}
We use the Adam optimizer in PyTorch to train the decision-maker, motion-planner, and goal-conditioned trajectory predictor. The decision-maker's learning rate starts at 5e-5 and decays to 1e-5 after 2000 gradient steps, while the motion-planner's learning rate starts at 1e-4 and also decays to 1e-5 after 2000 gradient steps. The learning rate for the goal-conditioned trajectory predictor is fixed at 1e-4. We train the framework using an AMD Ryzen 7 2700 CPU and an NVIDIA RTX 2080 Ti GPU.

\begin{figure}[!t]
    \centering
    \includegraphics[width=8.5cm]{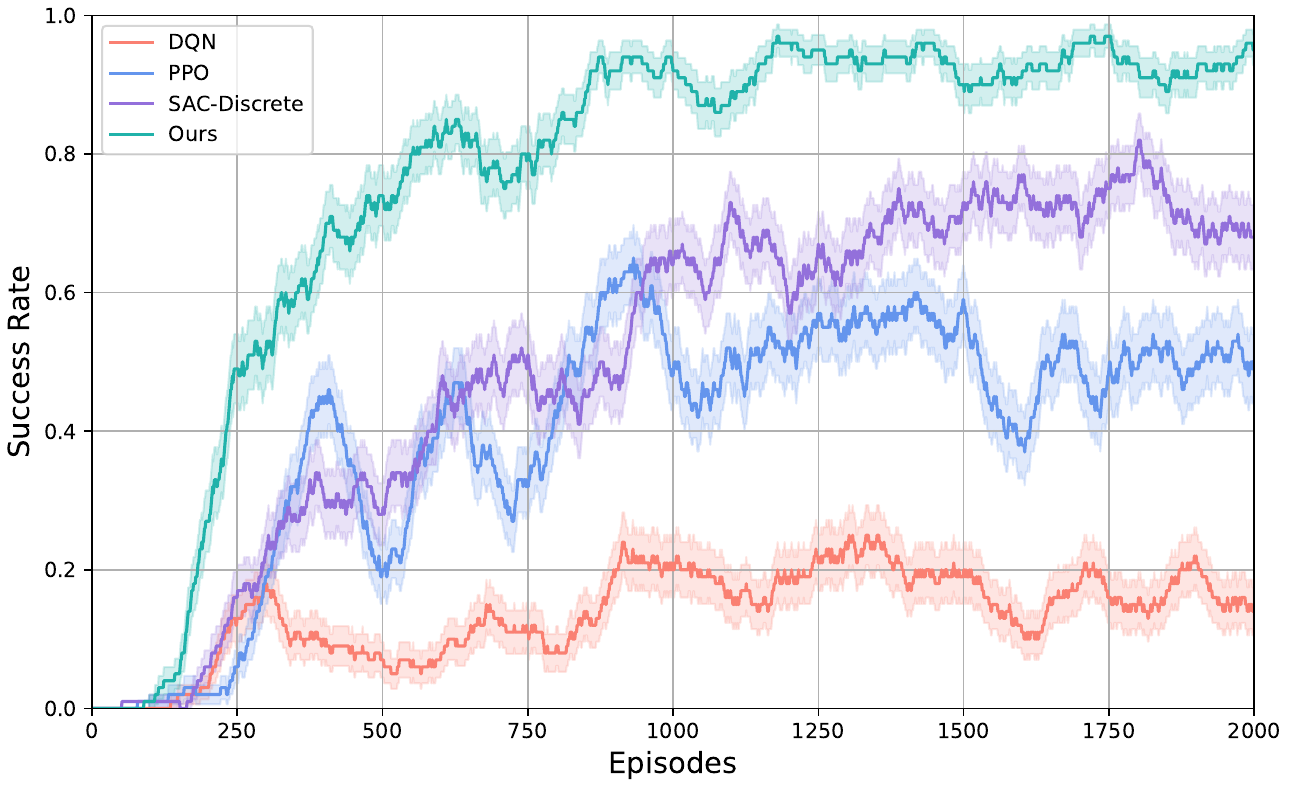}
    \caption{The training results of the proposed framework and baseline methods}
    \label{fig5}
\end{figure}

\begin{figure}[!t]
    \centering
    \includegraphics[width=8.5cm]{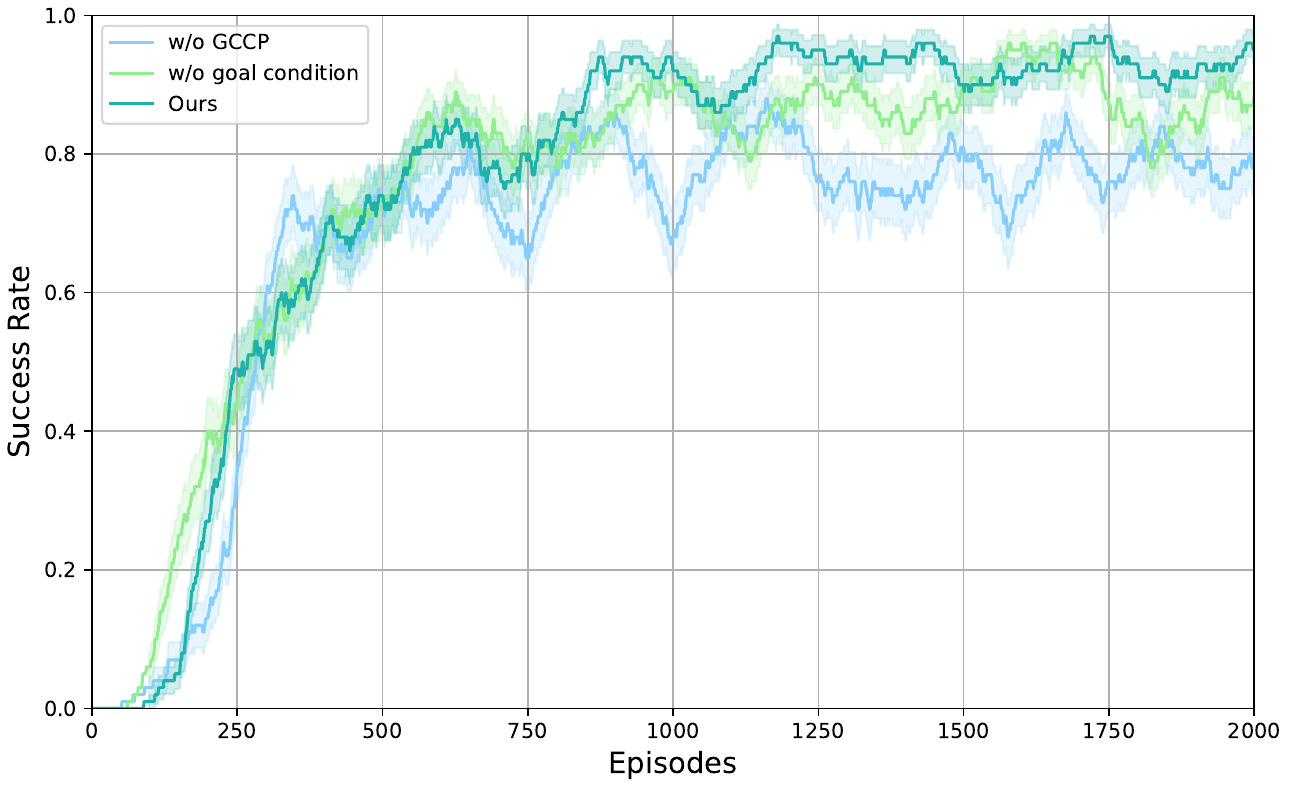}
    \caption{The training results of the proposed framework and ablated methods}
    \label{fig6}
\end{figure}

\begin{figure}[!t]
    \centering
    \includegraphics[width=8.5cm]{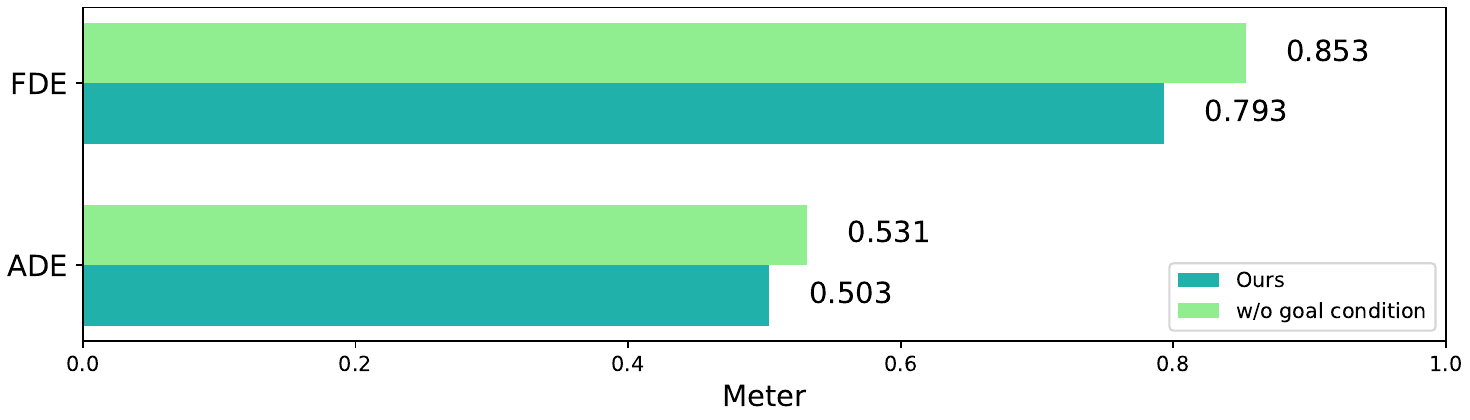}
    \caption{The trajectory prediction accuracy of the proposed framework and ablated method}
    \label{fig7}
\end{figure}

\begin{table}[]
\caption{TESTING RESULTS IN COMPARISON WITH RL METHODS}
\label{tab1}
\resizebox{\columnwidth}{!}{%
\begin{tabular}{|c|cccccc|cc|}
\hline
\multirow{2}{*}{Method} & \multicolumn{2}{c}{Turning Left} & \multicolumn{2}{c}{Going Straight} & \multicolumn{2}{c|}{Turning Right} & \multicolumn{2}{c|}{Overall}     \\
                        & Success         & Collision      & Success          & Collision       & Success          & Collision       & Success         & Collision      \\ \hline
DQN                     & 0\%             & 28\%           & 66\%             & 32\%            & 0\%              & 26\%            & 22.0\%          & 28.7\%         \\
PPO                     & 34\%            & 32\%           & 54\%             & 42\%            & 54\%             & 20\%            & 47.3\%          & 31.3\%         \\
SAC-Discrete            & 64\%            & 4\%            & 74\%             & 18\%            & 84\%             & 8\%             & 74.0\%          & 10.0\%         \\
Ours                    & \textbf{98\%}   & \textbf{2\%}   & \textbf{92\%}    & \textbf{6\%}    & \textbf{94\%}    & \textbf{2\%}    & \textbf{94.7\%} & \textbf{3.3\%} \\ \hline
\end{tabular}%
}
\end{table}

\begin{table}[]
\caption{TESTING RESULTS IN COMPARISON WITH ABLATED METHODS}
\label{tab2}
\resizebox{\columnwidth}{!}{%
\begin{tabular}{|c|cccccc|cc|}
\hline
\multirow{2}{*}{Method} & \multicolumn{2}{c}{Turning Left} & \multicolumn{2}{c}{Going Straight} & \multicolumn{2}{c|}{Turning Right} & \multicolumn{2}{c|}{Overall}     \\
                        & Success         & Collision      & Success          & Collision       & Success          & Collision       & Success         & Collision      \\ \hline
w/ CV                   & 86\%            & 10\%           & 88\%             & 12\%            & 90\%             & 4\%             & 88.0\%          & 8.7\%          \\
w/o GCCP                & 82\%            & 10\%           & 82\%             & 16\%            & 80\%             & 8\%             & 81.3\%          & 11.3\%         \\
w/o goal condition      & 94\%            & 2\%            & 86\%             & 8\%             & 94\%             & 4\%             & 91.3\%          & 4.7\%          \\
Ours                    & \textbf{98\%}   & \textbf{2\%}   & \textbf{92\%}    & \textbf{6\%}    & \textbf{94\%}    & \textbf{2\%}    & \textbf{94.7\%} & \textbf{3.3\%} \\ \hline
\end{tabular}%
}
\end{table}

\subsection{Results and Analysis}
\subsubsection{Training}
We train the framework in three intersection scenarios simultaneously. Fig.~\ref{fig5} illustrates the training process of the proposed framework compared to traditional RL methods, with the average success rate serving as the metric for decision-making and motion-planning. The results in Fig.~\ref{fig5} demonstrate that our method outperforms traditional RL methods in both sample efficiency and success rate. 
\textbf{The proposed method converges more quickly than traditional RL methods.} Specifically, the proposed method achieves a success rate exceeding 70\% by the 500th episode, whereas the success rates of traditional RL methods remain below 40\% at the same stage. Furthermore, by the end of training, the proposed method maintains a steady success rate of over 90\%, while traditional RL methods exhibit significantly lower performance, with none surpassing a success rate of 80\%. 
Fig.~\ref{fig6} illustrates the training process of the proposed framework alongside the ablated method. The results demonstrate that the proposed method incorporating the GCCP module achieves a higher success rate compared to the version without GCCP. Furthermore, the goal-conditioned version outperforms the non-goal-conditioned variant. 
For prediction accuracy, the evaluation metrics include the Average Displacement Error (ADE) and Final Displacement Error (FDE). As shown in Fig.~\ref{fig7}, the proposed method achieves superior prediction accuracy over the version without goal conditioning.

\subsubsection{Testing}
We evaluate the performance of our framework against baseline methods using 50 distinct traffic flows for each scenario. Table \ref{tab1} presents the testing results of our proposed method alongside several traditional reinforcement learning (RL) methods. The results demonstrate that our approach significantly outperforms traditional RL methods in terms of both success rate and collision rate. \textbf{Our proposed method is significantly safer than traditional RL methods}, with an overall collision rate of 3.3\%. In contrast, the collision rates of traditional RL methods all exceed 10\%, with DQN and PPO exhibiting collision rates as high as 30\%.
Table \ref{tab2} provides a comparison between our method and the ablated methods. The results indicate that the learning-based predictor outperforms the kinematic-based model. Additionally, the GCCP module not only significantly reduces the collision rate but also achieves a higher success rate.

\begin{figure}[!t]
    \centering
    \includegraphics[width=8.5cm]{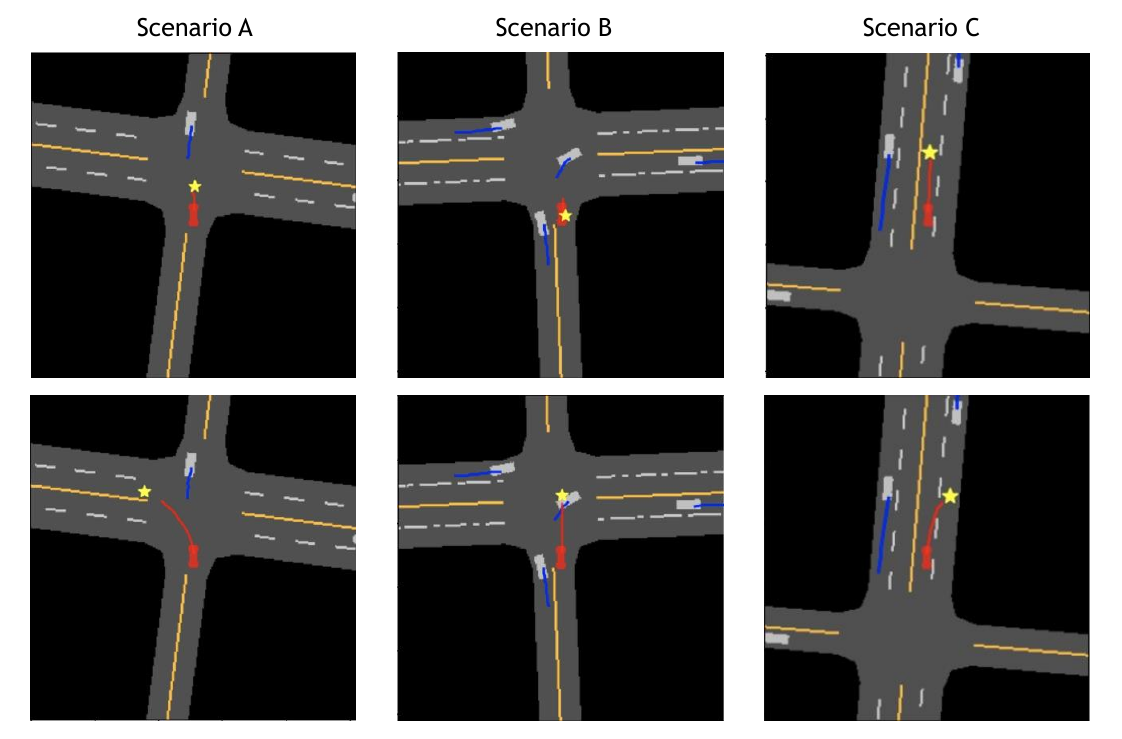}
    \caption{Three scenarios of the goal-conditioned prediction. The red rectangle represents the ego vehicle, while the white rectangles represent the surrounding vehicles. The yellow star indicates the selected subgoal for the ego vehicle. The red line shows the predicted trajectory of the ego vehicle, and the blue lines show the predicted trajectories of the surrounding vehicles. Each column corresponds to a scenario in which the ego vehicle is directed to perform one of two actions.}
    \label{fig8}
\end{figure}

\subsubsection{Goal-conditioned Prediction}
Goal-conditioned prediction enhances interpretability in the decision-making process.
Fig. \ref{fig8} shows three scenarios of the goal-conditioned prediction. In scenario A, the ego vehicle is directed to turn left, with an oncoming vehicle present. Two subgoals represent the ego vehicle's two possible actions: yielding or passing. In scenario B, the ego vehicle needs to stop to allow the oncoming vehicle to turn left. If the ego vehicle continues, it will collide with the oncoming vehicle. In Scenario C, the ego vehicle is instructed to either change lanes or remain in the same lane.

\section{CONCLUSIONS}
In this paper, we introduce a new HRL framework that incorporates the GCCP module. Within the hierarchical setup, the GCCP module assesses collision risks based on various potential subgoals. The high-level decision-maker selects the best subgoal, while the low-level motion-planner engages with the environment in alignment with the chosen subgoal. We train and evaluate the model using simulated intersection scenarios. The experimental results show that the proposed approach converges to an optimal policy more quickly and ensures greater safety compared to traditional RL methods.

% \addtolength{\textheight}{-12cm}   % This command serves to balance the column lengths
                                  % on the last page of the document manually. It shortens
                                  % the textheight of the last page by a suitable amount.
                                  % This command does not take effect until the next page
                                  % so it should come on the page before the last. Make
                                  % sure that you do not shorten the textheight too much.

%%%%%%%%%%%%%%%%%%%%%%%%%%%%%%%%%%%%%%%%%%%%%%%%%%%%%%%%%%%%%%%%%%%%%%%%%%%%%%%%

%%%%%%%%%%%%%%%%%%%%%%%%%%%%%%%%%%%%%%%%%%%%%%%%%%%%%%%%%%%%%%%%%%%%%%%%%%%%%%%%

%%%%%%%%%%%%%%%%%%%%%%%%%%%%%%%%%%%%%%%%%%%%%%%%%%%%%%%%%%%%%%%%%%%%%%%%%%%%%%%%
% \section*{APPENDIX}

% \section*{ACKNOWLEDGMENT}

%%%%%%%%%%%%%%%%%%%%%%%%%%%%%%%%%%%%%%%%%%%%%%%%%%%%%%%%%%%%%%%%%%%%%%%%%%%%%%%%

% \bibliographystyle{plain}
% 要引用的文献

% \bibliography{ref}

\bibliographystyle{IEEEtran}
\bibliography{IEEEabrv,ref}
% \bibliography{ref}

\end{document}